\documentclass{sig-alternate}
\usepackage{times}
\usepackage{helvet}
\usepackage{courier}
\usepackage[utf8]{inputenc}
\usepackage[T1]{fontenc}
\usepackage[linesnumbered,ruled]{algorithm2e}
\usepackage{color}
\usepackage{pgfplots}
\usepackage{amssymb}
\usepackage{pifont}
\usepackage{booktabs}
\usepackage{tabularx} 
\usepackage{algorithmicx,algpseudocode}
\usepackage{ifthen}
\usepackage{amsmath}
\usepackage{array}
\usepackage{tabu}

\usepackage{hyperref}
\usepackage{setspace}
\newtheorem{example}{Example}

\newcommand{\argmax}{\operatornamewithlimits{argmax}}

\begin{document}
\setlength{\pdfpagewidth}{8.5in}
\setlength{\pdfpageheight}{11in}

\CopyrightYear{2016} 
\conferenceinfo{3rd ACM SIGSPATIAL PhD Workshop'16,}{October 31-November 03 2016, Burlingame, CA, USA}

\title{Discovery of Driving Patterns by Trajectory Segmentation}

\numberofauthors{3} 
%

\author{Sobhan Moosavi, Arnab Nandi, and Rajiv Ramnath \\ \\
\affaddr{Department of Computer Science and Engineering}\\
\affaddr{Ohio State University}\\
\affaddr{Columbus, Ohio}\\
\email{\{moosavi.3,nandi.9,ramnath.6\}@osu.edu}
}

\maketitle

\begin{abstract}
Telematics data is becoming increasingly available due to the ubiquity of devices that collect data during drives, for different purposes, such as usage based insurance (UBI), fleet management, navigation of connected vehicles, etc. 
Consequently, a variety of data-analytic applications have become feasible that extract valuable insights from the data.
In this paper, we address the especially challenging problem of discovering behavior-based driving patterns from only externally observable phenomena (e.g. vehicle's speed).
We present a trajectory segmentation approach capable of discovering driving patterns as separate segments, based on the behavior of drivers. This segmentation approach includes a novel transformation of trajectories along with a dynamic programming approach for segmentation. 
We apply the segmentation approach on a real-word, rich dataset of personal car trajectories provided by a major insurance company based in Columbus, Ohio. 
Analysis and preliminary results show the applicability of approach for finding significant driving patterns. 
\end{abstract}



\section{Introduction}
\label{sec:intro}
The amount of telematics data has drastically increased thanks to the ubiquity of various types of devices and mobile apps to collect data during drive. Some instances of such transportation data are the New York taxi cab\footnote{\scriptsize http://toddwschneider.com/posts/analyzing-1-1-billion-nyc-taxi-and-uber-trips-with-a-vengeance/\smallskip} with 1.1 billion taxi trips and T-Drive \cite{yuan2010t} with trajectories of 10,357 Beijing taxi cabs for one week. 
Given the availability of these large transportation data sources, various analysis applications have been implemented to gain insights from this data.
Trajectory segmentation is one of the applications which tries into break a trajectory to several partitions or segments based on a set of optimization goals (e.g., minimizing the number of segments, maximizing homogeneity within segments, etc.), where each segment may represent a specific kind of movement pattern, phase, or behavior. 
In this paper, we propose a trajectory segmentation approach which is capable of discovering driving behavior patterns. 
Some examples of driving pattern are {\em make turn}, {\em change lane}, {\em merge highway}, etc.
We use the following example to describe the goal of current research in more detail. 

\begin{figure}
  \includegraphics[scale=0.28]{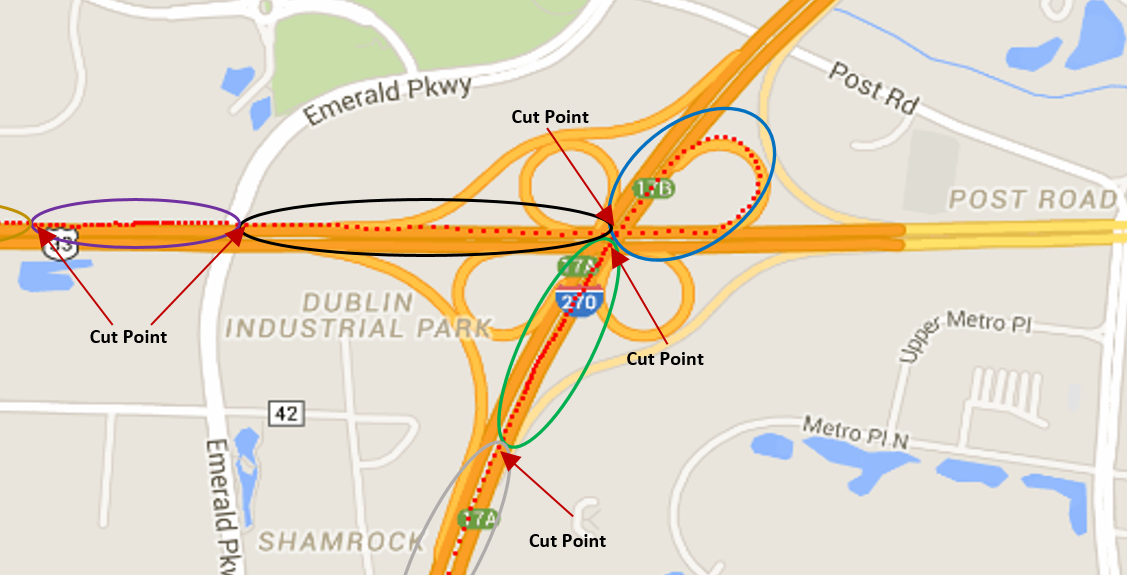}
  \caption{A sample trajectory with several behavior-based driving patterns specified by ovals. Transitions between patterns are pointed by arrows.}
  \label{fig:poc_traj}
\end{figure}

\begin{example}
\label{ex:segmentation}
Consider the trajectory in Figure~\ref{fig:poc_traj}. Red dots show the location of the car for every second of the trip. The trajectory begins at the bottom center and continues to the left after a clock-wise turn. Different parts of the trip exhibit different driving behavior-based patterns marked out by ovals. For instance, the green oval shows {\em slow movement}, where the captured locations are close to each other. Another pattern occurs when the car {\em enters the ramp and merges into a highway} (blue oval).
\end{example}

Example \ref{fig:poc_traj} is intended to illustrate that driving patterns are portions of a trajectory where there is homogeneity of driving behavior. 
The problem of finding significant driving patterns, as described by Example \ref{ex:segmentation}, is a challenging one for following reasons. 
First, unlike studies such as \cite{liu2001modeling,sathyanarayana2008driver} which collected data using a fully monitored environment (for example, with cameras placed inside the car monitoring the driver's every move and expression), and with a small set of drivers and routes, our dataset is the result of collecting data by observing only externally visible phenomena (e.g. vehicle's speed) with no additional intrusive monitoring. 
In addition, because of the size of the dataset of trajectories, and the potentially wide range of identifiable and useful driving patterns, a supervised approach is not viable. 
Thus, finding significant set of driving patterns is a challenging problem, worthy of our study. 

Discovery of behavior-based driving patterns is a part of a more generic framework for analysis of behavior of drivers to reveal how risky or safe are their driving habits. The result of such studies can be used for {\em usage based insurance}, {\em driver coaching}, {\em risk management}, and other related purposes. 
The main contribution of this paper is a novel trajectory segmentation approach to find driving patterns, based on the behavior of drivers. 
The rest of this paper is structured as follows: Section \ref{sec:problem_statement} provides the formal problem statement and required definitions. Detail of trajectory segmentation approach is addressed in Section \ref{sec:segmentation}. Next, the evaluation protocol and preliminary results are presented in Section \ref{sec:exp}. We provide a summary of related work in Section \ref{sec:rel}. Section \ref{sec:conc} concludes our study and describes potential future work.

\section{Problem Statement}
\label{sec:problem_statement}
Assume we are given a transportation database ${\mathcal D}$ of the form $\langle \Upsilon, \Gamma \rangle$ where $\Upsilon$ and $\Gamma$ are the set of vehicles and trajectories, respectively. 
Each trajectory $\gamma \in \Gamma$ is sequence of $|\gamma|$ data points $\langle \rho_1,\rho_2,\dots,\rho_{|\gamma|} \rangle$. Each data point $\rho$ is a tuple of the form $\{t,$ $lat,$ $lng,$ $s,$ $acc,$ $h\}$ which captures a vehicle's status at time $t$ as its latitude and longitude are $\langle lat, lng \rangle$, with speed $s$ (km/h), acceleration $acc$ ($m/s^2$), and heading $h$ (degrees). All time is assumed to be in seconds. Also, the heading is the direction of the moving vehicle, described by a degree-value between 0 and 359, where 0 means the north. 

A segmentation for a trajectory $\gamma$ into $n$ segments, denoted as $seg_{\gamma}$, is a set of cutting indexes $seg_{\gamma} = \langle I_1, I_2 \dots, I_n \rangle$ that mark the beginning points of the segments within a trajectory. Thus, we can define a set of cutting data points for the segmented trajectory ${\gamma}$ as $\langle p_{I_1}, p_{I_2} \dots, p_{I_n} \rangle$. Note that $p_{I_1}=\rho_1$. All data points between indexes $I_i$ and $I_{i+1}$, including point $\rho_{I_i}$ and excluding point $\rho_{I_{i+1}}$, belong to the $i^{th}$ segment. We denote the $i^{th}$ segment of $seg_{\gamma}$ as $seg_{\gamma}^{i}$ and its size as $|seg_{\gamma}^{i}|$. 
Note that segments are non-overlapping. Each segment represents a {\em driving pattern} and each cutting point $p_{I_i}, I_i \in seg_{\gamma}$, represents a {\em transition between patterns}. Figure \ref{fig:poc_traj} demonstrates segments (by ovals) and cutting points (by arrows) for a given trajectory. We define the optimization objectives for segmentation task as i) maximizing homogeneity within segments, ii) minimizing homogeneity between neighboring segments, and iii) minimizing the number of created segments.
\section{Segmentation approach}
\label{sec:segmentation}

We propose a novel approach to intelligently partition a trajectory, such that each resulting homogeneous segment corresponds to a specific driving pattern. Our trajectory segmentation approach includes following steps:
\begin{itemize}
  \item[i. ] Preprocessing of the trajectory dataset.
  \item[ii. ] Creating a memory-less Markov Model based on behavior of population of drivers in trajectory dataset.
  \item[iii. ] Using the Markov Model to transform a trajectory to a $signal$ in {\em Probabilistic Movement Dissimilarity (PMD)} space.
  \item[iv. ] Segmenting a signal by using a Dynamic Programming Segmentation approach and finding the best number of segments by Minimum Descriptor Length (MDL).
\end{itemize}
We next describe each step in more detail. 

\subsection{Preprocessing the Dataset}
Regarding the description of the data model in section \ref{sec:problem_statement}, the data set is a collection of trajectories, where each trajectory has a sequence of data points. The main steps for preprocessing the dataset are as follows:
    \begin{itemize}
      \item[--] Remove data points with missing or noisy (out of range) GPS records.
      \item[--] Normalize the values of $Acceleration$ and $Heading$ to be divisible by 0.25 and 5 respectively. This step helps to simplify the Markov Model, by reducing the number of possible states. 
      \item[--] Create training and test sets: We use the training set for creating the Markov Model and the test set for experiments.
    \end{itemize}

\subsection{Creating the Markov Model}
We create a memory-less Markov model $M = \{\Phi, \Delta, \Pi\}$, where $\Phi$ is the set of states, $\Delta$ is the set of transition between states (along with the frequency of each transition), and $\Pi$ is the set of probabilities of transition between the states. We use the following guidelines to create the $M$:
\begin{itemize}
      \item \underline{State}: We define a state $\phi \in \Phi$ as $\phi = \langle Speed,$ $Acceleration,$ $Heading \rangle$.
      \item \underline{Transition}: Given a trajectory $\gamma=\langle \rho_1, \rho_2, \dots, \rho_n \rangle$, for each pair of consecutive data points $\rho_i$ and $\rho_{i+1}$ of $\gamma$, where $1\leq i< n$, we create two states $\phi_i=\langle s_i, acc_i, h_i \rangle$ and $\phi_{i+1}=\langle s_{i+1}, acc_{i+1}, h_{i+1} \rangle$ for $\rho_i$ and $\rho_{i+1}$ respectively.
      We denote a transition from state $\phi_i$ to $\phi_{i+1}$ as $\phi_i \rightarrow \phi_{i+1}$.
      If $\Delta$ doesn't contain transition $\phi_i \rightarrow \phi_{i+1}$, then we insert $\langle \phi_i \rightarrow \phi_{i+1},1 \rangle$ into $\Delta$. Otherwise, we increase the frequency of transition $\phi_i \rightarrow \phi_{i+1}$ by 1.
      \item \underline{Probability of Transition}: For a specific state $\phi$, let us assume there is a $\delta\subseteq\Delta$ where $\delta=\{\langle \phi\rightarrow\phi_1,n_1\rangle, \dots, \langle \phi\rightarrow\phi_k,n_k\rangle\}$, and where $n_i$ is the number of observed transitions from $\phi$ to $\phi_i$ in the dataset, we update $\Pi$ by inserting the probability of each transition $\phi \rightarrow \phi_i$, $1\leq i\leq k$, using Equation \ref{eqn:trans_prob}:
      \begin{equation}\label{eqn:trans_prob}
        \small
        prob_{\phi \rightarrow \phi_i} = \frac{n_i}{\sum_{j=1}^{k} n_j}
      \end{equation}
\end{itemize}

\subsection{Transforming Trajectories}
The aim of our segmentation approach is to provide a segmentation of trajectories based on behavior of drivers. Hence, an important step is to transform an input trajectory to a signal in Probabilistic Movement Dissimilarity (PMD) space.
Suppose we have a trajectory $\gamma = \langle \rho_1, \rho_2, \dots, \rho_n \rangle$ and a Markov Model $M = \{\Phi, \Delta, \Pi\}$, we propose Algorithm~\ref{algo:transformation} to map $\gamma$ to a signal $S_{\gamma}$ in PMD space. 
Given consecutive data points $\rho_i,\rho_{i+1}\in\gamma$, Algorithm~\ref{algo:transformation} first maps them to states $\phi$ and $\phi'$ respectively. Then, it calculates how $unlikely$ is the transition $\phi\rightarrow\phi'$, based on $M$. 

\begin{algorithm}
\scriptsize
\DontPrintSemicolon
\KwIn{$\gamma$, $M$}
\KwOut{$S_{\gamma}$ \Comment{$S_{\gamma}$ is transformed version (signal) of $\gamma$}}
$S_{\gamma} \gets \langle\rangle$\;
\For {$i=1$  to  $n$-$1$}
{
    $\phi \gets ReturnState(M,\rho_i)$\;
    $\phi' \gets ReturnState(M,\rho_{i+1})$\;
    $v = 0$\;
    \If{$\phi\neq \phi'$}
    {
        $prob_{\phi\rightarrow \phi'} = ReturnProb(M,\phi,\phi')$\;
        $R \gets TransitionFrom(M, \phi)$\;
        \Comment{$R = \{r|$ $(\phi\rightarrow r)\in\Delta \}$}\;
        \For{$r \in R$}  
        {
            $prob_{\phi\rightarrow r} = ReturnProb(M,\phi,r)$\;
            $v$ += $Euclidean(\phi',r)\times prob_{\phi\rightarrow r}$\;
        }
        $v = \frac{v}{|R|}$\;
    }
    
    $S_{\gamma} \gets Append(S_{\gamma},v)$ \Comment{Appending $v$ at the end of $S_{\gamma}$}\;
}
\caption{{\small \tt Trajectory Transformation}}
\label{algo:transformation}
\end{algorithm}

In Algorithm~\ref{algo:transformation}, $ReturnState$ returns a state corresponding to input data point $\rho_i$, and $ReturnProb$ returns transition probability from $\phi$ to $\phi'$. $TransitionFrom$ returns a list of all states $r$ given an input state $\phi$, such that transition $(\phi \rightarrow r) \in \Delta$. Also, note that if $\phi$ and $\phi'$ represent the same state, then the transition is quite likely.
Based on this algorithm, we map a test trajectory to a signal in PMD space. The signal of a trajectory demonstrates the unlikelihood of behavior of driver during the trip. An unlikelihood score is calculated based on the transition probabilities in the Markov Model $M$. 
Lines 7 to 14 in Algorithm~\ref{algo:transformation} measure how far the observed transition $\phi\rightarrow\phi'$ is from our expectation regarding the Markov Model $M$.

\begin{figure}
  \includegraphics[scale=0.29]{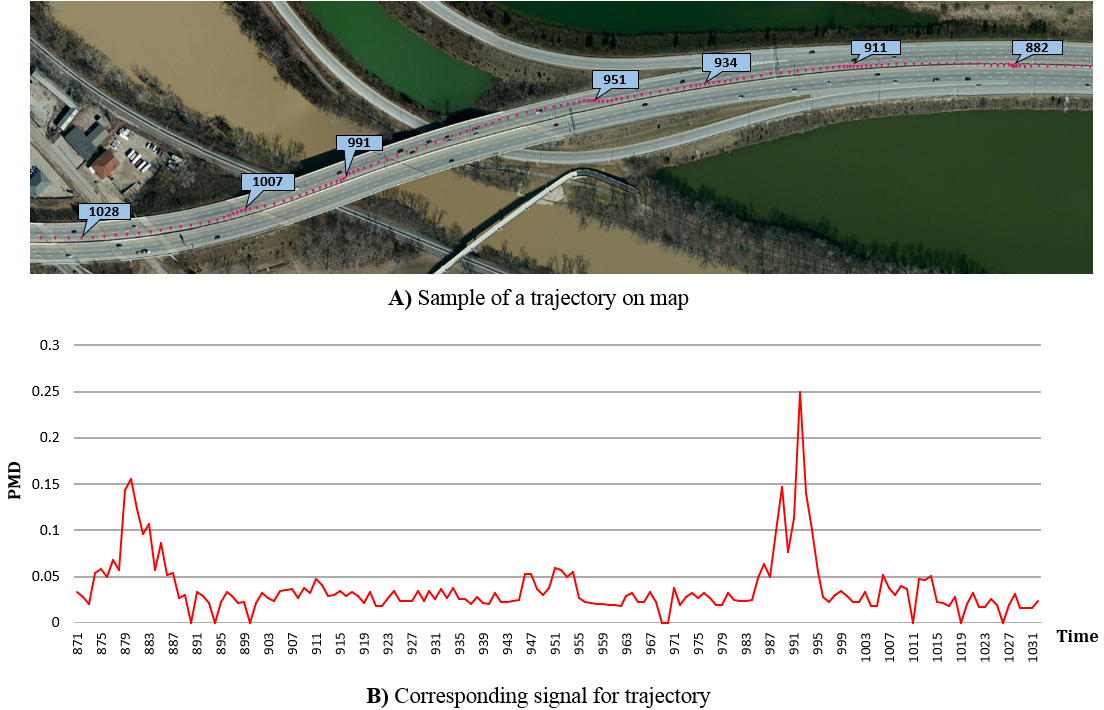}
  \caption{A) Sample trajectory on map with numbers in call-outs indicate timestamps B) The sample trajectory mapped to a signal in PMD space}
  \label{fig:pmd}
\end{figure}

Figure~\ref{fig:pmd} depicts a part of a sample trajectory and it's corresponding signal in PMD space. The numbers in rectangular call-outs in Figure~\ref{fig:pmd}.A show time stamps which can be matched with {\em Time} axis in Figure~\ref{fig:pmd}.B. The more unlikely the behavior of driver be, the larger the value of PMD is. For instance, a large PMD value is observable for time stamp 991 in Figure~\ref{fig:pmd}.B, where the actual trip in Figure~\ref{fig:pmd}.A shows an unexpected reduction in speed and also a lane change.

The main takeaway from this step is that we use a signal in PMD space as a representation of the behavior of a driver for a given trip, in comparison with the rest of the population of drivers and trajectories. 

\subsection{Dynamic Programming Trajectory Segmentation}
Once the signal for a trajectory has been created, the trajectory segmentation problem reduces to a {\em Signal Segmentation} problem. For segmenting a signal, we use an existing approach which has been successfully applied for segmenting electrical signals \cite{han2004optimal}. This approach is a dynamic programming algorithm that uses the Maximum Likelihood principle for segmenting one dimensional signals. 
Given an input signal $S = \langle x_1,x_2,\dots,x_N\rangle$, the Maximum Likelihood for $S$ can be defined by Equation \ref{eqn:ML}.
    \begin{equation}\label{eqn:ML}
        \small
        ML(\theta;x_1,x_{2},\dots,x_N) = f(x_1,x_{2},\dots,x_N|\theta)={\displaystyle \prod_{i=1}^{N} f(x_i|\theta)}
     \end{equation}
In this formula, $\theta$ is the set of parameters for a probability density function (PDF) $f$, which can be estimated based on data points of signal $S$. As in \cite{han2004optimal}, we leverage the {\em Gaussian distribution} to find the parameters of the PDF f, thus, $\theta = \langle\mu,\sigma \rangle$, where $\mu$ and $sigma$ are the sample mean and standard deviation respectively.  

Note that the goal of segmenting a trajectory $\gamma$ and it's signal $S_{\gamma} = \langle x_1,x_2,\dots,x_N\rangle$ (see section~\ref{sec:problem_statement}), is to find a set of cutting indexes $seg_{\gamma} = \langle I_1, I_2 \dots, I_n \rangle$, where $n\leq N$ is the best number of existing segments (i.e. with the greatest maximum likelihood). The recurrence relation for segmenting the signal $S_{\gamma}$ is defined below:
    \begin{equation}\label{eqn:recurrence}
        \small
        SSC(S_{\gamma}, 1, n) = \argmax_{2\leq i\leq N}(ML(S_{\gamma},1,i) + SSC(S_{\gamma}, i+1, n-1))
    \end{equation}
In Equation \ref{eqn:recurrence}, $SSC(S_{\gamma}, i, \nu)$ gives the best Segmentation Score for a sub-sequence of signal $S_{\gamma}$ which starts at index $i$, with the goal being to find $\nu$ segments. Also, $ML(S_{\gamma},i,j)$ gives the maximum likelihood score for sub-sequence $\langle x_i,x_{i+1},\dots,x_j\rangle$ of $S_{\gamma}$. Note that we assume the minimum length of a segment to be 2. More details of this algorithm may be found in \cite{han2004optimal}.

\begin{figure}
  \includegraphics[scale=0.34]{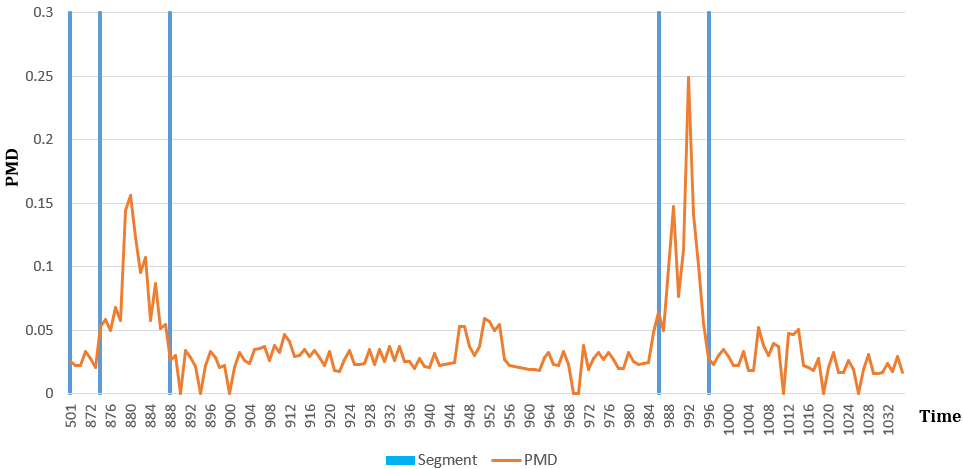}
  \caption{Segmentation of a sample trajectory, where the best number of segments is 5. One can observe the homogeneous pattern of behavior within each segment.}
  \label{fig:segmented_trip}
\end{figure}

The last question in this sub-section is: how to find the best number of existing segments within a signal? We use the Minimum Descriptor Length (MDL) \cite{rissanen1978modeling} for this purpose, which has been applied in \cite{han2004optimal} as well. MDL tries to minimize the Equation \ref{eqn:MDL} for $n=1,2,\dots,K$, where n is the number of segments and $K$ is the maximum possible number of segments (chosen by the user):
    \begin{equation}\label{eqn:MDL}
        \small
        MDL(n) = -ln{\displaystyle \prod_{i=1}^{n} f(x_{I_i},x_{I_i+1},\dots,x_{I_{i+1}-1},|\theta_i)} + \frac{r_n}{2}ln N
    \end{equation}
In Equation \ref{eqn:MDL}, $\theta_i$ is the parameter set of the corresponding PDF, $r_n$ is the number of estimated parameters (where $n$ is the number of segments), and $N$ is the length of the signal. Figure \ref{fig:segmented_trip} shows a part of a segmented signal which is related to the sample trajectory in figure~\ref{fig:pmd}.A. The blue lines in figure \ref{fig:segmented_trip} show the starting points of segments (i.e. the cutting points). The best number of segments which has been found by our MDL algorithm is 5. Note that we can observe the homogeneity of driving behavior patterns {\em within} segments and the heterogeneity of the driving patterns {\em between} segments.

As an example of driving behavior pattern which is captured by our trajectory segmentation approach, we point to the segment which starts at time stamp 986 in Figure \ref{fig:segmented_trip}. Regarding the actual trip in \ref{fig:pmd}.A, we see this segment is related to a part of driving behavior where driver reduces speed and changes the lanes.

\section{evaluation}
\label{sec:exp}
We first describe the dataset which is used in this study. Then, we provide experimental settings and some statistics as earlier results of trajectory segmentation approach which is applied on our real-world dataset \footnote{Code and sample data can be find in this GitHub repository: \url{https://github.com/sobhan-moosavi/Trajectory\_Segmentation}}.

\subsection{Trajectory Dataset}
We used a real-world dataset of 100,000 personal car trajectories provided by a major insurance company based in Columbus, Ohio. These trajectories were collected during 2011 to 2015.
We used approximately 95\% of trajectories for training (i.e. creating the Markov model) and 5\% as the test set (for evaluation).
The test dataset contains about 4,500 trajectories of 92 drivers for 5 different, popular routes in the city. Routes and number of trajectories for each is summarized in Table \ref{tab:segmentation_result}. 

\begin{table}
    \scriptsize
    \centering
    \caption{Summary of trajectory test set and segmentation result\\\hspace{\textwidth}}
    \begin{tabular}{c | c | c | c | c}
        \textbf{Route} & \textbf{\#Trajectories} & \textbf{Avg. Length} & \textbf{Avg. \#Segment}  & \textbf{Std. \#Segment} \\ [0.6ex]
        \hline
        315 Fwy & 426 & 705 & 8 & 7\\
        \hline
        I-270 & 701 & 389 & 4.9 & 3.8\\
        \hline
        I-670 & 443 & 392 & 7.4 & 6.4\\
        \hline
        I-70 & 1,572 & 324 & 5.4 & 4.9\\
        \hline
        I-71 & 1,320 & 549 & 7.5 & 6.8\\
    \end{tabular}
    \label{tab:segmentation_result}
\end{table}

\subsection{Segmentation results}
We used the process which is described in Section \ref{sec:segmentation} to segment trajectories in the test set. To find the the upper bound on the number of existing segments $K$ (Section \ref{sec:segmentation}), we used a heuristic as follows: for a given trajectory $\gamma$ of length $N$, we set $K=\frac{N}{10}$. Based on the segmentation result which is illustrated in Table \ref{tab:segmentation_result}, this is a reasonable upper bound. Note that the best number of segments is likely a result of the length of the trips in test set.
Table \ref{tab:segmentation_result} summarizes the segmentation results by providing the average and standard deviation for the number of segment for trajectories in different routes of the test set.
\section{Related Work}
\label{sec:rel}
Trajectory Segmentation, as described in Section \ref{sec:problem_statement}, has been addressed in the literature in several studies like \cite{buchin2010algorithmic,alewijnse2014framework, chen2013pathlet, anagnostopoulos2006global}. In \cite{buchin2010algorithmic}, a greedy segmentation algorithm exploits a set of monotonic spatio-temporal criteria (e.g., defining relative thresholds for some feature values) on features like speed, heading, etc. Alewijnse et al. extended the previous work to both monotonic and non-monotonic criteria \cite{alewijnse2014framework}. However, criteria-based methods need human input for tuning parameters. Moreover, they are {\em context-agnostic} in that they only consider the input trajectory and not the whole dataset. Therefore, the optimization process is a local one, where we propose a global optimization for segmentation. 

Our segmentation approach is a context-aware one by building a Markov Model for the whole dataset prior to segmentation. Similarly, some context-aware approaches are proposed in the literature including \cite{mann2002trajectory, alewijnse2014model}. Alewijnse et al. \cite{alewijnse2014model} present a context-aware approach which builds a Brownian Bridge model and uses a dynamic programming algorithm to capture the best set of segments of animal movements. While our solution bears some similarities with \cite{alewijnse2014model}, it exploits a normal distribution model instead, which we find it more suitable for car transportation data.

In \cite{panagiotakis2012segmentation}, a trajectory-to-signal transformation is performed prior to segmentation using similarity values between each line segment of input trajectory and the rest of the line segments in the dataset, using global voting. Then, segmentation discovery is done using a sliding-window approach. Our approach, in contrast, performs a behavior likelihood-based transformation to provide a behavior based segmentation and to find the segments which are representatives for driving behavior patterns. 
Essentially, our solution is a global optimization-based segmentation approach that builds up a model on the entire dataset. Note also that here is no need for human intervention in our solution as in \cite{buchin2010algorithmic,alewijnse2014framework}.
\section{Conclusion and future work}
\label{sec:conc}
In this paper, we proposed a Trajectory Segmentation approach to detect behavior based driving patterns for a given trajectory, based on externally observable phenomena. Our approach is a context aware solution which considers the behavior of the entire population of drivers to detect driving patterns. 
Our preliminary analysis based on existing use cases demonstrate the interpretability of segmentation results, as one of them described in Section \ref{sec:segmentation} for instance (Figures \ref{fig:pmd} and \ref{fig:segmented_trip}). 

We use the current study as a part of a more generic framework for analyzing the behavior of drivers to reveal how risky or safe their driving habits are. Other parts of this framework can be outlined as follows and they also will be considered as extensions of current study. 
In order to get more insight about extracted patterns by segmentation approach, we will design a supervised learning approach to learn and then predict true labels for patterns. Potential labels may be {\em making a turn}, {\em changing the lane}, {\em merging to a highway}, etc. 
Moreover, by having true labels for extracted patterns, we will apply sequential pattern mining techniques to extract significant sequences of driving patterns for a single driver or a population of drivers. Finally, by having human experts in the loop, we will identify the safe or risky sequences of driving patterns. In this way, we can formulate the problem of finding safe or risky drivers, based on their driving habits, as an end-to-end solution.

\bibliographystyle{abbrv}
\bibliography{main}

\begin{thebibliography}{10}

\bibitem{alewijnse2014framework}
S.~Alewijnse, K.~Buchin, M.~Buchin, A.~K{\"o}lzsch, H.~Kruckenberg, and M.~A.
  Westenberg.
\newblock A framework for trajectory segmentation by stable criteria.
\newblock In {\em Proceedings of the 22nd ACM SIGSPATIAL International
  Conference on Advances in Geographic Information Systems}, pages 351--360.
  ACM, 2014.

\bibitem{alewijnse2014model}
S.~P. Alewijnse, K.~Buchin, M.~Buchin, S.~Sijben, and M.~A. Westenberg.
\newblock Model-based segmentation and classification of trajectories.
\newblock In {\em Dead Sea, Israel: Proceedings of the 30th European Workshop
  on Computational Geometry March}, pages 3--5, 2014.

\bibitem{anagnostopoulos2006global}
A.~Anagnostopoulos, M.~Vlachos, M.~Hadjieleftheriou, E.~Keogh, and P.~S. Yu.
\newblock Global distance-based segmentation of trajectories.
\newblock In {\em Proceedings of the 12th ACM SIGKDD international conference
  on Knowledge discovery and data mining}, pages 34--43. ACM, 2006.

\bibitem{buchin2010algorithmic}
M.~Buchin, A.~Driemel, M.~van Kreveld, and V.~Sacrist{\'a}n.
\newblock An algorithmic framework for segmenting trajectories based on
  spatio-temporal criteria.
\newblock In {\em Proceedings of the 18th SIGSPATIAL International Conference
  on Advances in Geographic Information Systems}, pages 202--211. ACM, 2010.

\bibitem{chen2013pathlet}
C.~Chen, H.~Su, Q.~Huang, L.~Zhang, and L.~Guibas.
\newblock Pathlet learning for compressing and planning trajectories.
\newblock In {\em Proceedings of the 21st ACM SIGSPATIAL International
  Conference on Advances in Geographic Information Systems}, pages 392--395.
  ACM, 2013.

\bibitem{han2004optimal}
T.~X. Han, S.~Kay, and T.~S. Huang.
\newblock Optimal segmentation of signals and its application to image
  denoising and boundary feature extraction.
\newblock In {\em Image Processing, 2004. ICIP'04. 2004 International
  Conference on}, volume~4, pages 2693--2696. IEEE, 2004.

\bibitem{liu2001modeling}
A.~Liu and D.~Salvucci.
\newblock Modeling and prediction of human driver behavior.
\newblock In {\em Intl. Conference on HCI}, 2001.

\bibitem{mann2002trajectory}
R.~Mann, A.~D. Jepson, and T.~El-Maraghi.
\newblock Trajectory segmentation using dynamic programming.
\newblock In {\em Pattern Recognition, 2002. Proceedings. 16th International
  Conference on}, volume~1, pages 331--334. IEEE, 2002.

\bibitem{panagiotakis2012segmentation}
C.~Panagiotakis, N.~Pelekis, I.~Kopanakis, E.~Ramasso, and Y.~Theodoridis.
\newblock Segmentation and sampling of moving object trajectories based on
  representativeness.
\newblock {\em IEEE Transactions on Knowledge and Data Engineering},
  24(7):1328--1343, 2012.

\bibitem{rissanen1978modeling}
J.~Rissanen.
\newblock Modeling by shortest data description.
\newblock {\em Automatica}, 14(5):465--471, 1978.

\bibitem{sathyanarayana2008driver}
A.~Sathyanarayana, P.~Boyraz, and J.~H. Hansen.
\newblock Driver behavior analysis and route recognition by hidden markov
  models.
\newblock In {\em Vehicular Electronics and Safety, 2008. ICVES 2008. IEEE
  International Conference on}, pages 276--281. IEEE, 2008.

\bibitem{yuan2010t}
J.~Yuan, Y.~Zheng, C.~Zhang, W.~Xie, X.~Xie, G.~Sun, and Y.~Huang.
\newblock T-drive: driving directions based on taxi trajectories.
\newblock In {\em Proceedings of the 18th SIGSPATIAL International conference
  on advances in geographic information systems}, pages 99--108. ACM, 2010.

\end{thebibliography}

\end{document}